%% file: main.tex
\documentclass[10pt,twocolumn,letterpaper]{article}

\usepackage[pagenumbers]{cvpr} % To force page numbers, e.g. for an arXiv version

\input{preamble}

\definecolor{cvprblue}{rgb}{0.21,0.49,0.74}
\usepackage[pagebackref,breaklinks,colorlinks,allcolors=cvprblue]{hyperref}

\title{VLM-FO1: Bridging the Gap Between High-Level Reasoning and Fine-Grained Perception in VLMs}

\author{
 \textbf{Peng Liu\textsuperscript{1}},
 \textbf{Haozhan Shen\textsuperscript{3}},
 \textbf{Chunxin Fang\textsuperscript{2}},
 \textbf{Zhicheng Sun\textsuperscript{1}},
 \textbf{Jiajia Liao\textsuperscript{2}},
 \textbf{Tiancheng Zhao\textsuperscript{1,2,*}}
\\
 \textsuperscript{1}Om AI Research,
 \textsuperscript{2}Binjiang Institute of Zhejiang University,
 \\
  \textsuperscript{3}College of Computer Science and Technology, Zhejiang University
\\
  {\tt\small tianchez@zju-bj.com} 
}

\begin{document}
\maketitle
\input{sec/0_abstract}    
\input{sec/1_intro}
\input{sec/2_related_work}
\input{sec/3_methodology}
\input{sec/4_experiments}

\input{sec/5_ablation_studies}

\input{sec/6_conclusion}

{
    \small
    \bibliographystyle{ieeenat_fullname}
    \bibliography{main}
}

\input{sec/X_suppl}

\end{document}

%% file: preamble.tex
\usepackage{multirow}
\usepackage{makecell}
\usepackage{float}

%% file: sec/0_abstract.tex
\begin{abstract}
Vision-Language Models (VLMs) excel at high-level scene understanding but falter on fine-grained perception tasks requiring precise localization. This failure stems from a fundamental mismatch, as generating exact numerical coordinates is an challenge task for language-centric architectures. In this paper, we introduce VLM-FO1, a novel framework that overcomes this limitation by reframing object-centric perception from a brittle coordinate generation problem into a robust feature retrieval task. Our method operates as a plug-and-play module that integrates with any pre-trained VLM. It leverages a Hybrid Fine-grained Region Encoder (HFRE), featuring a Dual-Vision Encoder, to generate powerful region tokens rich in both semantic and spatial detail. A token-based referencing system then enables the LLM to seamlessly reason about and ground language in these specific visual regions. Experiments show that VLM-FO1 achieves state-of-the-art performance across a diverse suite of benchmarks, demonstrating exceptional capabilities in Object Grounding, Region Generative Understanding, and Visual Region Reasoning. Crucially, our two-stage training strategy ensures these perception gains are achieved without compromising the base model's general visual understanding capabilities. VLM-FO1 establishes an effective and flexible paradigm for building perception-aware VLMs, bridging the gap between high-level reasoning and fine-grained visual grounding.
\end{abstract}

%% file: sec/1_intro.tex
\section{Introduction}
\label{sec:intro}

The advent of Large Language Models (LLMs)~\cite{achiam2023gpt,touvron2023llama, peng2023instruction,bai2023qwen,team2023internlm,glm2024chatglm,dubey2024llama, yang2025qwen3,liu2024deepseek} has marked a paradigm shift in artificial intelligence, demonstrating profound capabilities in generation, reasoning, and instruction following. This success has been extended to the visual domain through Vision-Language Models (VLMs)~\cite{liu2023visual,liu2024improved,wang2024qwen2,chen2024internvl,zhu2025internvl3,hong2024cogvlm2,laurenccon2024matters,deitke2024molmo}, which integrate powerful vision backbones with LLMs to interpret and reason about visual content. By mapping visual features into the language model's embedding space, VLMs have achieved remarkable performance on high-level visual understanding tasks such as visual question answering (VQA) and image captioning. Further advancements, such as the application of reinforcement learning techniques like Group Relative Policy Optimization (GRPO)~\cite{shao2024deepseekmath}, have continued to enhance their complex reasoning abilities.

Despite these advances, a critical weakness persists: state-of-the-art VLMs struggle with fine-grained visual perception tasks that demand precise spatial localization, such as object detection and grounding. This deficiency severely limits their applicability in real-world scenarios like autonomous robotics, detailed image analysis, and human-computer interaction, where understanding what is in an image is inseparable from knowing where it is. Our evaluations reveal a stark performance gap. On standard benchmarks like COCO~\cite{lin2014microsoft}, specialized detection models routinely achieve a mean Average Precision (mAP) of 50-60. In contrast, even a leading open-source model like Qwen2.5-VL-72B~\cite{bai2025qwen2} achieves a recall of less than 40\%, indicating a fundamental inability to reliably locate all relevant object instances.

This limitation stems from a core architectural mismatch: generating precise numerical coordinates is an "unnatural" task for models fundamentally designed for sequential language generation~\cite{karthick2025LLMReg}. The requirement to produce a string of exact floating-point numbers in a specific format is brittle; a single incorrect token can render an entire bounding box prediction invalid. This problem is exacerbated in scenes with multiple instances, where the generation of a long, structured sequence of coordinates challenges the model's attention mechanism, leading to low recall and compounding errors.

\begin{figure*}[t]
  \centering
  \includegraphics[height=16cm, width=0.8\textwidth]{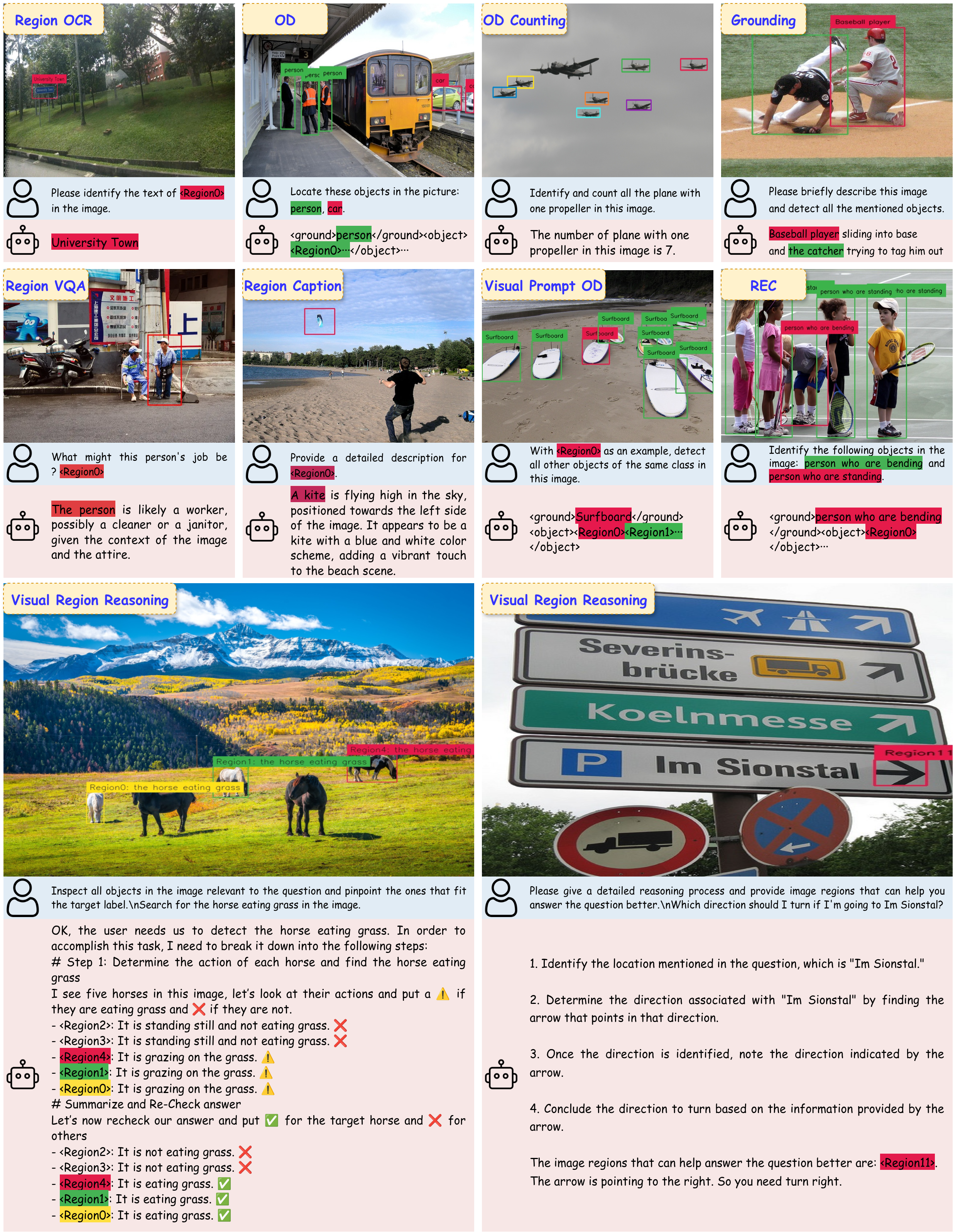}
  \caption{Visualization of VLM-FO1's perception abilities on diverse visual tasks.}
  \label{fig:fo1_structure}
\end{figure*}

To address this perception weakness, several approaches have been explored. Some methods~\cite{chen2021pix2seq,xiao2024florence,wang2022ofa} quantize object coordinates into a discrete vocabulary, simplifying the generation task. However, this approach still struggles with multiple instances and suffers from quantization errors, particularly on high-resolution images. Another strategy~\cite{lai2024lisa,pi2023detgpt,zhang2024llava} involves appending external prediction heads to the VLM to handle localization. While potentially effective, these modules introduce additional latency and often require complex, task-specific loss functions for optimization. Recently, another direction has emerged that employs agentic, vision-level reasoning. Methods like ZoomEye~\cite{shen2024zoomeye} treat the image as a navigable structure, allowing the VLM to dynamically "zoom in" on specific regions to gather fine-grained visual evidence.

A more promising direction~\cite{ma2024groma,jiang2024chatrex,cheng20253d} reframes the problem by using an external detection model to generate region proposals, effectively converting the difficult generation task into a simpler retrieval task. While innovative, existing methods in this vein have significant drawbacks. They either require joint end-to-end training with the detection model, creating a monolithic and cumbersome system, or they necessitate training a new architecture from scratch on massive datasets. Crucially, both approaches fail to leverage the rich visual understanding and world knowledge already embedded within large-scale, pre-trained VLMs, effectively discarding a powerful and readily available resource.

In response to these challenges, we introduce VLM-FO1, a novel framework that endows pre-trained VLMs with superior fine-grained perception without compromising their inherent strengths. The core idea is simple: we shift the paradigm from generating box coordinates to directly perceiving the content within them. VLM-FO1 treats any bounding box as a "visual prompt," extracts its features, and converts them into distinct "region tokens" that are fed directly into the LLM. This elegantly transforms object detection into a simple and accurate retrieval task.

Our primary innovations are threefold. First, VLM-FO1 is designed as a plug-and-play enhancement module that can be integrated with any pre-trained VLM, preserving its original capabilities while dramatically improving perception. Second, we introduce a novel Hybrid Fine-grained Region Encoder (HFRE), which features a Dual-Vision Encoder structure. This combines the VLM's original semantic-rich vision encoder with a new perception-enhanced vision encoder, yielding powerful object features that capture both high-level meaning and fine-grained detail. Third, our two-stage decoupled framework separates the training of the proposal model and the VLM. This modularity grants users the flexibility to pair VLM-FO1 with any proposal detector best suited for their specific application.

When combined with our custom-trained Omni Proposal Network (OPN), our lightweight VLM-FO1-3B model achieves 44.4 mAP on COCO, an improvement of over 20 points that places it on par with specialized detectors and far ahead of other VLMs. This strong performance extends to a wide variety of other region-related perception tasks, such as Referring Expression Comprehension (REC), object counting, and OCR, demonstrating the versatility and effectiveness of our approach. In summary, our main contributions are:
\begin{itemize}
\item A Flexible and Modular Plug-and-Play Framework: We propose VLM-FO1, a perception enhancement framework whose two-stage, decoupled design allows it to be seamlessly integrated with any pre-trained VLM. This modularity enables practitioners to use off-the-shelf object detectors for proposal generation, enhancing fine-grained perception without requiring full retraining or compromising the VLM's original capabilities.
\item A Novel Hybrid Fine-grained Region Encoder (HFRE): We introduce a Dual-Vision Encoder architecture that combines a semantic-rich vision encoder with a perception-enhanced tower to produce region tokens that are rich in both high-level meaning and fine-grained spatial detail.
\item State-of-the-Art Performance: We demonstrate the effectiveness of our method by achieving state-of-the-art results across a diverse suite of benchmarks spanning three key perspectives: Object Grounding, Region Generative Understanding, and Visual Region Reasoning, setting a new standard for perception-enhanced VLMs.
\end{itemize}

%% file: sec/2_related_work.tex
\section{Related Work}
\label{sec:formatting}

\subsection{Vision-Language Models (VLMs)}

Since the emergence of large language models (LLMs)\cite{chung2024scaling,dubey2024llama,touvron2023llama2,bai2023qwen}, they have achieved remarkable success across a wide range of linguistic applications, which has in turn fostered the development of Vision-Language Models (VLMs). Early pioneering works include \cite{alayrac2022flamingo,li2023blip,koh2023grounding}. Building on these foundations, LLaVA\cite{liu2023visual} leveraged GPT-4~\cite{achiam2023gpt} to construct training data, achieving strong performance in visual dialogue and reasoning, and inspiring a line of research on visual instruction data~\cite{liu2024improved,instructblip,chen2023sharegpt4v}. A typical architecture of VLMs encodes visual information through a vision encoder~\cite{pham2021learning,zhai2023sigmoid,EVA-CLIP} and integrates the resulting visual tokens with textual tokens within the LLM backbone. Today, some of the most widely adopted open-source VLM families include LLaVA\cite{liu2023visual,liu2024llavanext,li2024llava}, QwenVL\cite{wang2024qwen2,bai2025qwen2,Qwen-VL}, and InternVL~\cite{chen2024expanding,chen2024internvl,zhu2025internvl3,wang2025internvl3}.

\begin{figure*}[t]
  \centering
  \includegraphics[height=9cm, width=1.0\textwidth]{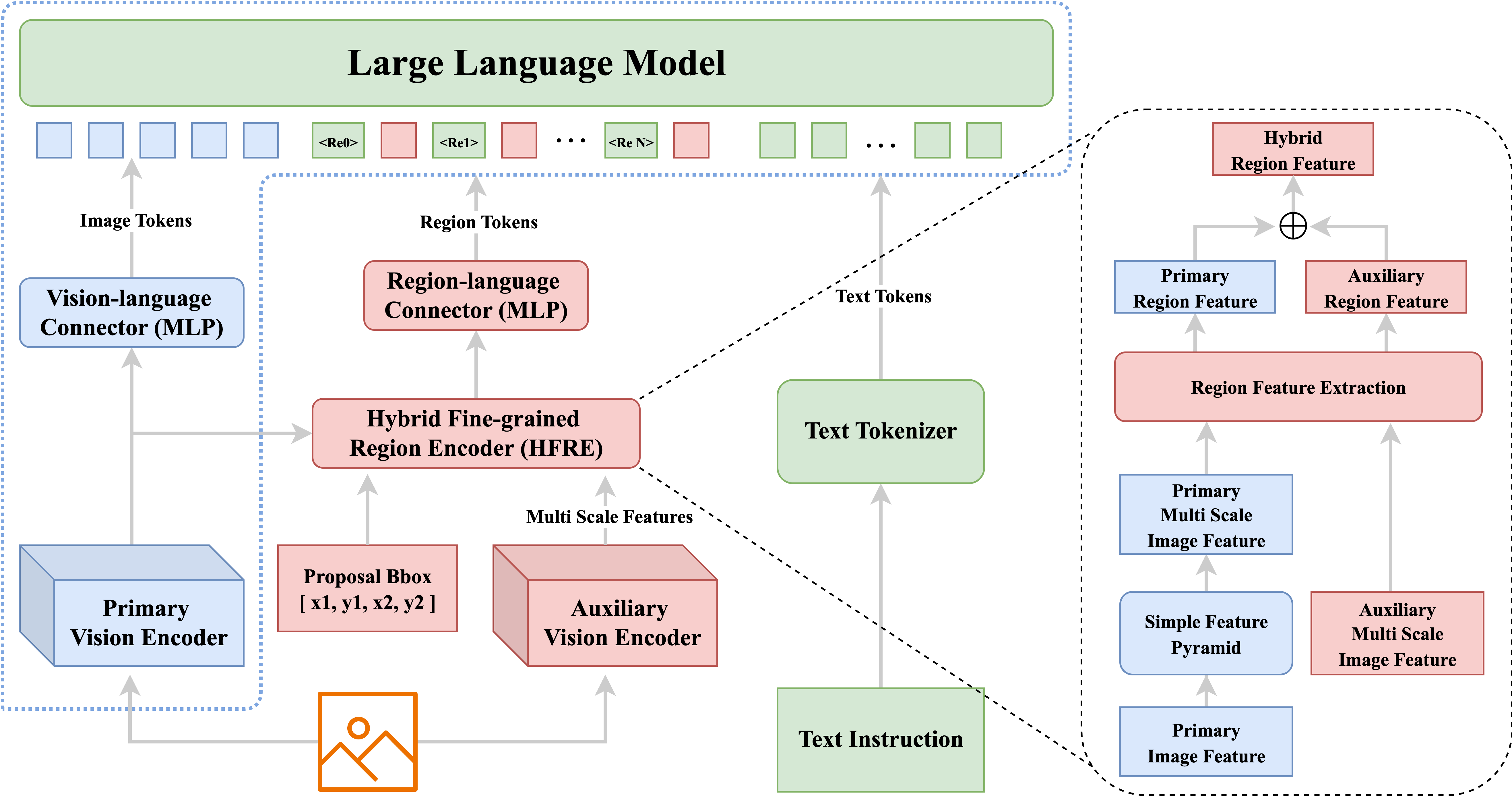}
  \caption{\textbf{An overview of our proposed model architecture.} The components enclosed within the blue dotted line represent a standard pretrained VLM, which can be initialized with existing weights to preserve its original performance. Our method introduces external modules, including a dual-vision encoder system and a Hybrid Fine-grained Region Encoder (HFRE), to enrich the VLM's fine-grained perception. These modules process an image to extract and fuse multi-scale region features, which are then fed as new region tokens to the VLM. The inset on the right details the generation of the Hybrid Region Feature.}
  \label{fig:fo1_structure}
\end{figure*}

\subsection{VLMs with Detection Enhancement}

To equip VLMs with detection capability, prior works have explored integrating detection heads~\cite{vstar,wu2024visionllm,lai2024lisa} or incorporating visual expert models~\cite{jiao2024lumen,shen2023hugginggpt,zhao2023bubogpt,ma2024groma}. More generally, most approaches adopt an auto-regressive strategy to sequentially generate the four coordinates of bounding boxes, well aligned with LLM backbones~\cite{you2023ferret,chen2023shikra,peng2023kosmos}. Building on this paradigm, the Griffon series~\cite{zhan2025griffonv1,zhan2024griffon-2,zhan2024griffon-g} progressively unified localization tasks, introduced high-resolution perception structures, and curated multi-dimensional datasets, extending VLMs to both vision-language and vision-centric settings. More recently, general-purpose VLMs~\cite{bai2025qwen2,wang2024qwen2,chen2024internvl,zhu2025internvl3,wang2025internvl3} trained with detection data have demonstrated strong performance, and reinforcement learning methods such as GRPO~\cite{shao2024deepseekmath} have been employed to further enhance visual reasoning for precise detection~\cite{shen2025vlm,liu2025visual}.

% Similar to our work, Chat-Rex\cite{jiang2024chatrex} treats detection as region-token retrieval by extracting region features and feeding them into the LLM. Different from it, our VLM-FO1 (1) employs a plug-and-play Hybrid Fine-Grained Region Encoder (Fig.\ref{fig:fo1_structure}), which produces more representative features and aligns better with the original visual feature space, whereas Chat-Rex change the original VLM architecture, and (2) is able to distinguish negative sample categories specified in the input prompt (i.e., categories that are not actually present in the image) when performing OD task, whereas Chat-Rex does not possess this capability.

Several prior works, notably Groma~\cite{ma2024groma} and ChatRex~\cite{jiang2024chatrex}, share a similar high-level approach by treating detection as a region-token retrieval task with extra detector. However, these methods typically necessitate significant architectural modifications to the base VLM or require joint training with the object detector, mandating a costly retraining process from scratch. In contrast, our VLM-FO1 employs a plug-and-play Hybrid Fine-Grained Region Encoder that seamlessly integrates with existing pre-trained VLMs, preserving their powerful, pre-learned representations while producing more representative and better-aligned region features. Furthermore, a critical distinction lies in the handling of negative categories. While prior models are often limited to detecting only the positive instances present in an image, our training strategy enables VLM-FO1 to robustly distinguish and reject categories specified in the prompt that are not actually present, a crucial capability for real-world applications.

\subsection{VLMs with Visual Prompt}  

Visual prompts take multiple forms. \textbf{Bounding-box prompts} provide coarse localization, as in Shikra~\cite{chen2023shikra} and InstructDET~\cite{dang2023instructdet}, which directly encode user-specified regions for object grounding. \textbf{Marker-based prompts}, such as Set-of-Mark (SoM)~\cite{yang2023set} and its extension SoM-LLaVA~\cite{yan2024list}, overlay symbolic cues (e.g., circles, arrows, IDs) to highlight target regions. ViP-LLaVA~\cite{cai2024vipllava} extends this idea by supporting arbitrary free-form markers like scribbles and arrows.  

\textbf{Pixel-level prompts} enable more fine-grained perception. OMG-LLaVA~\cite{OMGLLaVA} and VisionLLM~\cite{wang2023visionllmlargelanguagemodel} tokenize pixel-centric features, while DOrA~\cite{wu2024dora} and CoLLaVO~\cite{lee2024collavo} integrate pixel-level annotations to enhance semantic grounding. ControlMLLM~\cite{wu2024controlmllm} further leverages training-free visual prompt learning to model semantic pixel-text relations.  

In addition, \textbf{soft visual prompts} provide learnable perturbations in the pixel space. Transferable Visual Prompting (TVP)~\cite{Zhang_2024_CVPR} explores prompt generalization across MLLMs, while BlackVIP~\cite{Oh_2023_CVPR} and ILM-VP~\cite{Chen_2023_CVPR} design task-specific perturbations to adapt pre-trained models efficiently.  

%% file: sec/3_methodology.tex
\section{Methodology}

\subsection{Model Architecture}
The overall architecture of VLM-FO1, depicted in Figure~\ref{fig:fo1_structure}, is designed as a set of plug-and-play modules that augment a standard pre-trained VLM. Instead of altering the core VLM or training from scratch, our framework introduces specialized components to process region-level visual information. Given an image and a set of bounding box proposals, our model extracts and fuses multi-scale region features, projects them into "region tokens," and feeds them alongside global image and text tokens into the Large Language Model. This allows the LLM to perform reasoning grounded in specific, fine-grained visual evidence. The key components of our architecture are the Dual-Vision Encoder and the Hybrid Fine-grained Region Encoder (HFRE).

\textbf{Proposal Regions.} A core tenet of our VLM-FO1 framework is its two-stage, decoupled design, which makes the generation of proposal regions entirely independent from the VLM's perception module. This separation facilitates independent training and optimization of each component. More importantly, it provides exceptional flexibility, allowing users to switch between different object detectors to generate proposal regions based on specific scenarios, or even manually input regions of interest, all without any additional training. For our experiments and training, we developed an Omni Proposal Network (OPN), a variant based on OmDet-Turbo~\cite{zhao2024real}, to serve as a general-purpose detector. The OPN is trained to identify all potential foreground objects in an image, and we use it to generate the proposal regions for all our training and evaluation data.

\textbf{Dual-Vision Encoder.} The foundation of our fine-grained understanding capability is the Dual-Vision Encoder, a system engineered to produce region features that are simultaneously rich in semantic meaning and perceptual detail. It synergistically combines two components: the VLM's original Primary Vision Encoder and a new Auxiliary Vision Encoder. The primary encoder, having been co-trained with the LLM, excels at generating semantically-aligned features but lacks spatial precision due to its training on lower-resolution, global scenes. To compensate, the auxiliary encoder acts as a high-resolution detail specialist, processing the image at a higher fidelity to extract multi-scale feature maps rich in the perceptual cues (e.g., edges, textures) necessary for precise localization. While the primary encoder continues to provide global image tokens and semantic context for regions, the auxiliary encoder supplies the critical fine-grained information. The outputs from both are then intelligently fused to create a superior hybrid region feature representation.

\textbf{Hybrid Fine-grained Region Encoder (HFRE).} The HFRE is responsible for processing the multi-scale features from the Dual-Vision Encoder and generating the final region tokens for the LLM. This process involves three main stages: multi-scale feature extraction, hybrid feature fusion, and tokenization.

For the auxiliary vision encoder, we select DaViT~\cite{ding2022davit}, a vision transformer that combines a CNN-like multi-scale architecture with an efficient dual attention mechanism to capture both fine-grained local details and long-range global context. We extract a set of feature maps, $\{A_1, A_2, A_3, A_4\}$, from its backbone. These maps are up-sampled via interpolation to match the spatial dimensions of the largest feature map and then concatenated along the channel dimension to form a dense feature map $A_{\mathit{concat}}$. Given N proposal bounding boxes $B=b_1,...,b_N$, we use RoIAlign~\cite{he2017mask} followed by mean pooling to extract the corresponding region features, yielding the auxiliary region features $F_{\mathit{aux}} \in \mathbb{R}^{N \times D_a}$.

For the ViT-based primary vision encoder, which lacks a native feature pyramid, we introduce a Simple Feature Pyramid (SimpleFP) module, inspired by ViTDet~\cite{li2022exploring}. This module takes the last feature map from the encoder and applies a series of convolutions and deconvolutions with strides $\{2, 1, 1/2, 1/4\}$ to construct a feature pyramid. Similar to the auxiliary process, we then use RoIAlign to extract region features, resulting in the primary region features $F_{\mathit{pri}} \in \mathbb{R}^{N \times D_p}$.

Finally, the two sets of features are fused by concatenation to form a combined feature representation, $F_{\mathit{comb}} = \text{Concat}(F_{\mathit{pri}}, F_{\mathit{aux}}) \in \mathbb{R}^{N \times (D_p+D_a)}$. To explicitly provide the model with spatial information, we compute sine-cosine positional embeddings $E_{\mathit{pos}}$ from the coordinates of the proposal boxes and add them to the combined features: $F_{\mathit{hybrid}} = F_{\mathit{comb}} + E_{\mathit{pos}}$. This hybrid feature is then passed through a Region-Language Connector, an MLP layer, which projects it into the LLM's embedding space to produce the final region tokens.

\subsection{Grounding Language to Regions via Token-based Referencing}
To ground language in specific visual regions, our framework, inspired by previous works~\cite{jiang2024chatrex,ma2024groma}, establishes a token-based referencing system that enriches the LLM's input with explicit, addressable region-level information. We augment the standard VLM input of image and text tokens by introducing our new region tokens. To enable the LLM to distinguish between and reference specific regions, we introduce a set of N special tokens, \texttt{<region0>, <region1>, ..., <regionN-1>}, which serve as unique region index tokens.

The input to the LLM is structured as an interleaved sequence where each region token is preceded by its corresponding index token. This results in a final format of: \path{<image_tokens>\n<region0><region_token>...<regionN-1><region_token>\n<text_tokens>}. This structure allows the LLM to directly associate the region features with a unique identifier. Consequently, a user can refer to specific regions within a text prompt by simply using the corresponding index token.
%, for example: "What is the relationship between \texttt{<region1>} and \texttt{<region5>}?"

For the model's output, we introduce special tokens to handle grounding tasks. The \texttt{<ground></ground>} tokens demarcate a noun phrase in the response that requires grounding, and the \texttt{<object></object>} tokens enclose the region index tokens that correspond to that phrase. For instance, a valid grounded response would be: ``The \texttt{<ground>}people\path{</ground><object><region2><region10></object>} are dancing.'' This format provides an unambiguous link between textual concepts and their visual referents. For tasks that require simpler referencing without explicit grounding, the model can directly refer the region index token within its natural language response. This structured output format transforms complex localization into a native referencing task for the LLM.

%%%%%%%%%%%%%%%%%%%%%%%%%%%%%%%%%%%%%%%%%%%%%%%%%%%%%%%%%%%%%%
\begin{table*}[htbp]
\small
\centering
\begin{tabular}{l|l|l|c|c|c}
\toprule
\textbf{Type} & \textbf{Sub-type} & \textbf{Model} & \textbf{MSCOCO val2017} & \textbf{ODinW13} & \textbf{OVDEval} \\
\midrule
% --- Detection Model 分组 (总共6行) ---
\multirow{6}{*}{\textbf{\makecell{Detection\\Model}}} & \multirow{3}{*}{\textbf{OD}} & \textbf{Faster RCNN}~\cite{ren2015faster} & 42.0 & - & - \\
 & & \textbf{DETR}~\cite{carion2020end} & 43.3 & - & - \\
 & & \textbf{DINO}~\cite{zhang2022dino} & 49.4 & - & - \\
 \cmidrule(l){2-6}
 & \multirow{3}{*}{\textbf{OVD}} & \textbf{GLIP}~\cite{li2022grounded} & 49.8 & 52.1 & 18.4 \\
 & & \textbf{Grounding DINO}~\cite{liu2024grounding} & 52.5 & 55.7 & 25.3 \\
 & & \textbf{OmDet-Turbo}\cite{zhao2024real} & 53.4 & 54.1 & 25.9 \\
\midrule
% --- VLM 分组 (现在总共12行) ---
\multirow{12}{*}{\textbf{VLM}} & \multirow{2}{*}{\textbf{Close-source}} & \textbf{Gemini 1.5 Pro}~\cite{team2024gemini} & - & 36.7\textsuperscript{*} & - \\
 & & \textbf{GPT-4o}~\cite{hurst2024gpt} & 3.1 & - & - \\
\cmidrule(l){2-6}
% Open-source
 & \multirow{4}{*}{\textbf{Open-source}} & \textbf{InternVL2.5-8B}~\cite{chen2024expanding} & 12.1 & 20.2\textsuperscript{*} & - \\
 & & \textbf{InternVL2.5-72B} & - & 31.7\textsuperscript{*} & - \\
 & & \textbf{Qwen2.5-VL-7B}~\cite{bai2025qwen2} & 17.7 & 37.3\textsuperscript{*} & - \\
 & & \textbf{Qwen2.5-VL-72B} & - & 43.1\textsuperscript{*} & - \\
\cmidrule(l){2-6}
% OD-enhanced
 & \multirow{6}{*}{\textbf{OBJ-enhanced}} & \textbf{VLM-R1-7B}~\cite{shen2025vlm} & - & - & 31.0 \\
 & & \textbf{Lumen}~\cite{jiao2024lumen} & 35.3 & - & - \\
 & & \textbf{Griffon v2}~\cite{zhan2024griffon-2} & 38.5 & - & - \\
 & & \textbf{Griffon-G-7B}~\cite{zhan2024griffon-g} & 40.2 & 43.8\textsuperscript{*} & - \\
 & & \textbf{ChatRex-7B}~\cite{jiang2024chatrex} & 4.3(48.2 reported) & - & - \\
 \cmidrule(l){3-6}
 & & \textbf{VLM-FO1-3B(Ours)} & \textbf{44.4} & \textbf{44.0} & \textbf{43.7} \\
\bottomrule
\end{tabular}
\caption{\textbf{Object Grounding performance on OD benchmarks.} * indicates evaluation under a simplified setting where only ground-truth categories are queried.} 
\label{tab:object_grounding_benchmark}
\end{table*}
%%%%%%%%%%%%%%%%%%%%%%%%%%%%%%%%%%%%%%%%%%%%%%%%%%%%%%%%%%%%%%

\subsection{Training Strategy}
The training of VLM-FO1 is conducted in two distinct stages designed to efficiently integrate fine-grained perception while preserving the model's extensive pre-trained knowledge.

\textbf{Stage 1: Region-Language Alignment Training.} The primary objective of this initial stage is to align the newly introduced region tokens with the LLM's feature space with minimal disruption to the existing VLM weights. To achieve this, we first extend the LLM's vocabulary with our special tokens (e.g., \texttt{<RegionN>}, \texttt{<ground>}) and freeze the embeddings of the original vocabulary, ensuring that only the new token embeddings are updated. Concurrently, we freeze the parameters of the entire pre-trained VLM, including the primary vision encoder and the LLM itself. Training is focused exclusively on the newly added modules: the HFRE and the Region-Language Connector. This isolated training strategy allows the model to learn a robust mapping from visual regions to token space.

\textbf{Stage 2: Perception Instruction Finetuning.} The second stage aims to holistically enhance the model's perception capabilities by fine-tuning the integrated system on a broader set of instruction-based tasks. In this phase, we unfreeze the parameters of the Auxiliary Vision Encoder, the HFRE, the Region-Language Connector, and the LLM. The Primary Vision Encoder remains frozen throughout the entire training process, acting as a stable anchor for the original VLM's semantic understanding. The training dataset is expanded to include a wider variety of perception-focused instruction data. 

%% file: sec/4_experiments.tex
\section{Experiments}

\subsection{Experimental Setup}
\textbf{Model Setup.} Our experiments are built upon the Qwen2.5-VL model, chosen for its excellent baseline performance in visual understanding. For the auxiliary vision encoder, we integrate a pre-trained DaViT-Large model. Within the HFRE module, the auxiliary encoder extracts features from 4 multi-scale layers, resulting in a region feature of dimension 3840. The primary vision encoder utilizes the SimpleFP module to generate 4 multi-scale features, each with a dimension of 512, which are then combined into a 2048-dimension feature. The final hybrid region feature thus has a dimension of 5888. For our two-stage training, we set the learning rate to 1e-3 for Stage 1 and 1e-5 for Stage 2. For each image, we process a maximum of 100 input proposals, selecting the top 100 predictions from our OPN based on their confidence scores.

\textbf{Training Data.} Our training data is structured to support our two-stage training strategy, as summarized in Table~\ref{tab:training_data}.
\begin{itemize}
    \item Stage 1 (Region-Language Alignment): In this stage, training is focused on aligning the visual features from the HFRE with the LLM's embedding space. To achieve this, we use a curated collection of datasets centered on region-language tasks. This includes large-scale object detection datasets (COCO~\cite{lin2014microsoft}, O365~\cite{shao2019objects365}, V3Det~\cite{wang2023v3det}), grounding data (GOLDG~\cite{kamath2021mdetr}), and region caption data (Rexverse-2M~\cite{jiang2024chatrex}).
    \item Stage 2 (Perception SFT): The second stage broadens the model's capabilities by training on a diverse mix of perception-focused instruction datasets. In addition to the data from Stage 1, we incorporate datasets for REC, grounding, region captioning, region reasoning, counting, region QA, and OCR. Furthermore, for detection-related tasks, we introduce rejection samples for 20\% of the data, where the model is prompted to find objects that are not present in the image. This strategy encourages the model to be more discerning and avoid hallucinating objects based solely on the text prompt. Crucially, to mitigate catastrophic forgetting, we also include a subset of data from the OmChat-SFT collection (which contains data from LLaVA-1.5~\cite{liu2024improved}, The Cauldron\cite{laurenccon2024matters}, CogVLM\cite{hong2024cogvlm2}, etc.). This mix of conventional VLM task data ensures that the model retains its high-level scene interpretation abilities while mastering fine-grained perception.
\end{itemize}

%---------------------------------------------
\begin{table}
\centering
\begin{tabular}{l cc|cc}
\toprule
% --- 表头第一行 ---
\multirow{2}{*}{\textbf{Model}} & \multicolumn{2}{c|}{\textbf{LVIS}} & \multicolumn{2}{c}{\textbf{PACO}} \\
% --- 表头第二行 ---
\cmidrule(lr){2-3} \cmidrule(lr){4-5}
& \textbf{SS} & \textbf{S-IoU} & \textbf{SS} & \textbf{S-IoU} \\
\midrule
% --- 数据行 ---
LLaVA-1.5~\cite{liu2023visual} & 49.0 & 19.8 & 42.2 & 14.6 \\
Kosmos-2~\cite{peng2023kosmos} & 39.0 & 8.7 & 32.1 & 4.8 \\
Shikra-7B~\cite{chen2023shikra} & 49.7 & 19.8 & 43.6 & 11.4 \\
GPT4RoI-7B & 51.3 & 12.0 & 48.0 & 12.1 \\
Ferret-7B~\cite{you2023ferret} & 63.8 & 36.6 & 58.7 & 26.0 \\
Osprey-7B~\cite{yuan2024osprey} & 65.2 & 38.2 & 73.1 & 52.7 \\
VisionLLM v2-7B~\cite{wu2024visionllm} & 68.9 & 46.3 & 67.7 & 44.0 \\
VP-SPHINX-13B~\cite{lin2024draw} & 87.1 & 62.9 & 76.8 & 51.3 \\
DAM-8B~\cite{lian2025describe} & 89.0 & 77.7 & 84.2 & 73.2 \\
PAM-3B~\cite{lin2025perceive} & 88.6 & 78.3 & 87.4 & 74.9 \\
ChatRex-7B~\cite{jiang2024chatrex} & 89.8 & 82.6 & \textbf{91.4} & \textbf{85.1} \\
\midrule
\textbf{VLM-FO1-3B (Ours)} & \textbf{92.4} & \textbf{86.4} & 88.1 & 77.6 \\
\bottomrule
\end{tabular}
\caption{\textbf{Region-level classification performance of VLMs on LVIS and PACO datasets.}}
\label{tab:lvis_paco_benchmark}
\end{table}

\subsection{Main Results}
To comprehensively assess the effectiveness of VLM-FO1, we assess its capabilities across three key dimensions: Object Grounding, Region Generative Understanding, and Visual Region Reasoning. We benchmark our model on a diverse suite of tasks within each of these areas.

%---------------------------------------------
\begin{table}
\centering
\begin{tabular}{l | c}
\toprule
% --- 表头第一行 ---
\textbf{Model} & \textbf{\makecell{COCO Text\\Accuracy(\%)}} \\
\midrule
% --- 数据行 ---
ChatSpot-7B~\cite{zhao2023chatspot} & 31.8 \\
PAM-3B~\cite{lin2025perceive} & 42.2 \\
VP-LLAVA-8B~\cite{lin2024draw} & 44.8 \\
VP-SPHINX-13B~\cite{lin2024draw} & 45.4 \\
\midrule
\textbf{VLM-FO1-3B (Ours)} & \textbf{59.0} \\
\bottomrule
\end{tabular}
\caption{\textbf{Regional OCR performance on COCOText benchmark.}}
\label{tab:cocotext_benchmark}
\end{table}

%---------------------------------------------

%---------------------------------------------
\begin{table}[h!]
\centering
\begin{tabular}{l | c}
\toprule
% --- 表头第一行 ---
\textbf{Model} & \textbf{\makecell{Ferret Bench\\Refer. Reasoning}} \\
\midrule
% --- 数据行 ---
LLaVA-7B~\cite{liu2023visual} & 31.7 \\
Kosmos-2~\cite{peng2023kosmos} & 33.7 \\
Osprey-7B~\cite{yuan2024osprey} & 67.8 \\
Ferret-13B~\cite{you2023ferret} & 68.7 \\
Ferret-v2-13B~\cite{zhang2024ferret} & 79.4 \\
VP-LLAVA-8B~\cite{lin2024draw} & 68.9 \\
VP-SPHINX-13B~\cite{lin2024draw} & 71.4 \\
\midrule
\textbf{VLM-FO1-3B (Ours)} & \textbf{80.1} \\
\bottomrule
\end{tabular}
\caption{\textbf{Referring Reasoning performance of Ferret Bench.}}
\label{tab:ferret_benchmark}
\end{table}
%---------------------------------------------

\textbf{Object Grounding.} We first evaluate the model's core ability to ground language in objects through detection tasks. As shown in Table~\ref{tab:object_grounding_benchmark}, we benchmark VLM-FO1 on standard object detection with COCO~\cite{lin2014microsoft}, open-vocabulary detection in real-world settings with ODinW13~\cite{li2022elevater}, and challenging language-based detection with hard negatives on OVDEval~\cite{yao2024evaluate}.

The results clearly demonstrate VLM-FO1's superiority. On COCO and ODinW13, VLM-FO1 significantly outperforms other VLM-based models, showcasing its powerful perception and high recall. For instance, GPT-4o~\cite{hurst2024gpt} achieves a mere 3.1 mAP on COCO, confirming that even the most advanced models fail at direct coordinate regression. While ChatRex-7B~\cite{jiang2024chatrex} reports a high mAP of 48.2, this is achieved under a non-standard evaluation protocol where only the ground-truth categories for each image are provided as queries; under standard COCO evaluation, its performance drops to 4.3 mAP, likely due to an inability to handle negative categories. VLM-FO1 successfully overcomes both of these fundamental challenges.

More impressively, on ODinW13, our model achieves the highest score despite being tested under the rigorous, standard mAP protocol. It is important to note that many other VLMs (marked with *) are evaluated on ODinW13 using a simplified setting where only ground-truth categories are fed to the model individually. This easier setting avoids the challenge of distinguishing hard negatives. Even against models tested in this simplified manner, VLM-FO1, under standard evaluation, still comes out on top.

%---------------------------------------------
\begin{table*}[h]
\centering
\begin{tabular}{l ccc ccc cc ccc}
\toprule
% --- 表头第一行 ---
\multirow{2}{*}{\textbf{Model}} & \multicolumn{3}{c}{\textbf{Refcoco}} & \multicolumn{3}{c}{\textbf{Refcoco+}} & \multicolumn{2}{c}{\textbf{Refcocog}} & \multicolumn{3}{c}{\textbf{HumanRef}}\\
% --- 表头第二行 ---
\cmidrule(lr){2-4} \cmidrule(lr){5-7} \cmidrule(lr){8-9} \cmidrule(lr){10-12}
& val & testA & testB & val & testA & testB & val & test & P & R & DF1\\
\midrule
% --- 数据行 ---
Gemini 1.5 Pro~\cite{team2024gemini} & 73.2 & 72.9 & 74.6 & 62.5 & 63.9 & 65.0 & 75.2 & 76.2 & - & - & - \\
DINOX~\cite{ren2024dino} & - & - & - & - & - & - & - & - & 33.1 & 75.2 & 23.3 \\
Grounding DINO~\cite{liu2024grounding} & 90.6 & 93.2 & 88.2 & 88.2 & 89.0 & 75.9 & 86.1 & 87.0 & 33.1 & 75.2 & 23.3 \\
InternVL-2.5-8B~\cite{chen2024expanding} & 90.3 & 94.5 & 85.9 & 85.2 & 91.5 & 78.8 & 86.7 & 87.6 & 37.8 & 29.8 & 31.9 \\
Ferret-7B~\cite{you2023ferret} & 87.5 & 91.4 & 82.5 & 80.8 & 87.4 & 73.1 & 83.9 & 84.8 & 43.2 & 34.4 & 34.3 \\
Groma-7B~\cite{ma2024groma} & 89.5 & 92.1 & 86.3 & 83.9 & 88.9 & 78.1 & 86.37 & 87.01 & 48.7 & 65.9 & 42.1 \\
ChatRex-7B~\cite{jiang2024chatrex} & 91.0 & \textbf{94.1} & 87.0 & \textbf{89.8} & 91.9 & 79.3 & \textbf{89.8} & \textbf{90.0} & 72.2 & 50.4 & 55.6 \\
Qwen2.5-VL-7B~\cite{bai2025qwen2} & 90.0 & 92.5 & 85.4 & 84.2 & 89.1 & 76.9 & 87.2 & 87.2 & 68.5 & 52.5 & 56.2 \\
Molmo-7B-D~\cite{deitke2024molmo} & - & - & - & - & - & - & - & - & 82.5 & 77.7 & 72.6 \\
RexSeek-7B~\cite{jiang2025referring} & - & - & - & - & - & - & 84.0 & 84.4 & 85.8 & 85.9 & 82.3 \\
% Rex-Thinker-GRPO & - & - & - & - & - & - & - & - & 86.8 & 86.6 & 83.5 \\
\midrule
\textbf{VLM-FO1-3B(Ours)} & \textbf{91.12} & 93.7 & \textbf{87.6} & 86.4 & \textbf{91.9} & \textbf{80.6} & 88.9 & 88.3 & 87.1 & 83.3 & \textbf{82.6} \\
\bottomrule
\end{tabular}
\caption{\textbf{Model Performance on Referring Benchmarks}}
\label{tab:rec_benchmark}
\end{table*}
%---------------------------------------------

%---------------------------------------------
\begin{table}[ht!]
\small
\centering
\begin{tabular}{p{0.68cm} p{3.6cm} | c | c}
\toprule
\textbf{Type} & \textbf{Model} & \textbf{\makecell{CountBench\\Acc(\%)}} & \textbf{\makecell{PixMo\\Count}}\\
\midrule
% --- Close-source models 分组 (跨5行) ---
\multirow{5}{*}{\makecell{Close\\Source}} & GPT-4V~\cite{2023GPT4VisionSC} & 69.9 & 45.0 \\
& GPT-4o-0513~\cite{hurst2024gpt} & 87.9 & 59.6 \\
& Gemini 1.5 Pro~\cite{team2024gemini} & 85.8 & 64.3 \\
& Claude-3 Opus~\cite{anthropic2024claude} & 83.6 & 43.3 \\
& Claude-3.5 Sonnet~\cite{anthropic2024claude} & 89.7 & 58.3 \\
\midrule
% --- Open-source models 分组 (跨9行) ---
\multirow{9}{*}{\makecell{Open\\Source}} & LLaVA-1.5-13B~\cite{liu2024improved} & 47.1 & 35.2 \\
& LLaVA OneVision-72B~\cite{li2024llava} & 84.3 & 60.7 \\
& InternVL2-8B~\cite{chen2024expanding} & 57.8 & 43.9 \\
& InternVL2-Llama-3-76B~\cite{chen2024expanding} & 74.7 & 54.6 \\
& InternVL2.5-78B~\cite{chen2024expanding} & 72.1 & - \\
& Qwen2-VL-72B~\cite{wang2024qwen2} & 80.4 & 55.7 \\
%& Qwen2.5-VL-72B & \textbf{93.6} & - \\
& Pixtral-12B~\cite{agrawal2024pixtral} & 78.8 & 51.7 \\
& Llama-3.2V-90B-Instruct~\cite{dubey2024llama} & 78.5 & 58.5 \\
& Molmo-7B-D~\cite{deitke2024molmo} & 88.5 & 84.8 \\
& Molmo-72B~\cite{deitke2024molmo} & \textbf{91.2} & 85.2 \\
\midrule
% --- VLM-FO1-3B 单独一行 ---
& \textbf{VLM-FO1-3B (Ours)} & 87.8 & \textbf{86.0}\\
\bottomrule
\end{tabular}
\caption{\textbf{Model Performance on Object Counting Benchmarks}}
\label{tab:counting_benchmark}
\end{table}
%---------------------------------------------

The most compelling results are on OVDEval, which evaluates performance on linguistic labels with hard negatives. Here, VLM-FO1 surpasses not only other VLMs but also specialized detection models like Grounding DINO. This highlights a key advantage of our approach: VLM-FO1 effectively leverages the world knowledge, entity recognition, and reasoning abilities inherited from its VLM foundation to disambiguate complex and challenging text prompts.

\textbf{Region Generative Understanding.} We further evaluate our model's ability to understand and generate accurate textual descriptions based on specific visual regions. This is tested across three diverse tasks: region-level classification (Table~\ref{tab:lvis_paco_benchmark}), region-based OCR (Table~\ref{tab:cocotext_benchmark}), and referring reasoning (Table~\ref{tab:ferret_benchmark}). The results unequivocally demonstrate VLM-FO1's SOTA capabilities. On the object-level LVIS and part-level PACO~\cite{ramanathan2023paco} datasets, our model sets a new state-of-the-art for region classification, with our efficient 3B model outperforming significantly larger 8B and 13B models. Our architecture demonstrates a strong capability for generating precise text targeting fine-grained regions. On the COCOText~\cite{veit2016coco} benchmark for regional OCR, VLM-FO1 achieves a staggering 59.0\% accuracy, surpassing the next best model by over 13 points. Finally, on the challenging referring reasoning subset of Ferret Bench~\cite{you2023ferret}, our model achieves a new SOTA score of 80.1, demonstrating that its strong fine-grained perception directly translates to a more accurate understanding of specific visual regions and their relationships.

\begin{table*}[h!]
%\footnotesize
\small
\begin{tabular}{l|c|c|c|c|c|c|c|c|c}
\toprule
\textbf{Model} &\textbf{AVG} & \textbf{\makecell{MMBench\\v1.1~\cite{liu2024mmbench}}} & \textbf{\makecell{AI2D\\~\cite{kembhavi2016diagram}}} &\textbf{\makecell{MMStar\\~\cite{chen2024we}}} & \textbf{\makecell{Hallusion\\Bench\\~\cite{guan2024hallusionbench}}} & \textbf{\makecell{OCR\\Bench\\~\cite{liu2024ocrbench}}} & \textbf{\makecell{MathVista\\~\cite{lu2023mathvista}}} & \textbf{\makecell{MMVet\\~\cite{lu2023mathvista}}} & \textbf{\makecell{MMMU\\Val\\~\cite{yue2024mmmu}}} \\
& &  & & & & & & & \\
\midrule
\textbf{Qwen2.5-VL-3B} & 64.5 & 76.8 & 81.4 & 56.3 & 46.6 & 82.8 & 61.2 & 60 & 51.2 \\
\midrule
\textbf{VLM-FO1-3B} & \textbf{64.6} & 78.2 & 81.2 & 56.9 & 47.9 & 82.3 & 65.6 & 54.9 & 49.9 \\
\bottomrule
\end{tabular}
\caption{\textbf{Comparison of general VLM capabilities on OpenCompass benchmarks.}}
\label{tab:opencompass_benchmark}
\end{table*}

\begin{table}[htbp]
  \small
  \centering
  \begin{tabular}{l c}
    \toprule
    \textbf{Model}                                      & \textbf{Average Score} \\
    \midrule
    VLM-FO1-3B                                          & \textbf{67.65} \\
    VLM-FO1-3B (QwenViT unfrozen)                       & 66.35 \\
    \midrule
    Only Aux. Region Feat. (DaViT unfrozen)           & 65.89 \\
    Only Prim. Region Feat. (QwenViT frozen)             & 65.76 \\
    Only Prim. Region Feat. (QwenViT unfrozen)           & 66.15 \\
    \bottomrule
  \end{tabular}
  \caption{\textbf{Ablation study for HFRE module.} (Aux. Region Feat. denotes Auxiliary Region Feature, and Prim. Region Feat. denotes Primary Region Feature.) }
  \label{tab:HFRE_ablation}
\end{table}

\textbf{Visual Region Reasoning.} In this section, we evaluate the model's ability to leverage its fine-grained region features to perform complex reasoning. We benchmark this on two challenging task families: Referring Expression Comprehension (Table~\ref{tab:rec_benchmark}) and Object Counting (Table~\ref{tab:counting_benchmark}). In REC tasks such as Refcoco/+/g, the model must reason about a natural language description to identify the correct object. Our model achieves consistently top-tier results across these benchmarks. More significantly, on HumanRef~\cite{jiang2025referring}, a difficult benchmark focusing on people with complex descriptions (attributes, positions, interactions) and hard negatives, VLM-FO1 demonstrates remarkable performance, achieving a new state-of-the-art. This success underscores its robust reasoning capability and its skill in disambiguating between visually similar instances based on nuanced language. Furthermore, in Object Counting, a task notorious for causing failures in large VLMs, our model excels by adopting a "Detect-then-Count" reasoning process. It first localizes all target instances and then aggregates the count, leading to superior accuracy on CountBench~\cite{paiss2023teaching} and the challenging PixMo-Count~\cite{deitke2024molmo} benchmark. This methodical approach allows our compact V-FO1-3B to outperform even much larger closed-source models like GPT-4V and open-source models up to 72B parameters, highlighting the power of grounding complex reasoning in accurate, fine-grained perception.

\subsection{Preservation of General VLM Capabilities}
A critical aspect of our framework is its ability to enhance fine-grained perception without degrading the base model's general visual understanding capabilities. To validate this, we evaluated our VLM-FO1-3B model against the original Qwen2.5-VL-3B on the comprehensive OpenCompass~\cite{2023opencompass} benchmark. The results, shown in Table~\ref{tab:opencompass_benchmark}, confirm that our method successfully avoids catastrophic forgetting.

Across a wide range of general VLM benchmarks—including MMBench~\cite{liu2024mmbench}, AI2D~\cite{kembhavi2016diagram}, and MMStar~\cite{chen2024we}—our VLM-FO1 model's performance remains virtually identical to the base Qwen2.5-VL model, with a negligible difference in the average score. This demonstrates the effectiveness of our two-stage training strategy; by initially freezing the core VLM during the Region-Language Alignment phase and subsequently mixing in general VLM instruction data during the second stage, our method successfully prevents the degradation of pre-existing knowledge. The results confirm that VLM-FO1 acts as a true enhancement module, allowing pre-trained VLMs to gain superior fine-grained perception while retaining their powerful and general visual understanding abilities.

%% file: sec/5_ablation_studies.tex
\begin{table}[htbp]
  \centering
  \begin{tabular}{l c}
    \toprule
    \textbf{Model}                  & \textbf{Average Score} \\
    \midrule
    \makecell[l]{Only Prim. Region Feat.\\with SimpleFP}          & \textbf{66.15} \\
    \midrule
    \makecell[l]{Only Prim. Region Feat.\\ w/o SimpleFP}   & 64.94 \\
    \bottomrule
  \end{tabular}
  \caption{\textbf{Impact of SimpleFP on Primary Region Feature Performance.}}
  \label{tab:simplefp_ablation}
\end{table}

\section{Ablation Studies}
To validate the effectiveness of the individual components in our model design, we conduct a series of ablation studies. For efficiency, all ablation experiments are conducted by training on a representative subset of our full perception SFT dataset. The ``Average Score" reported in this section is an average calculated from the benchmark scores evaluated in the previous sections.

\subsection{Effectiveness of Components in HFRE}
We first analyze the contribution of the different components within the HFRE module. As shown in Table~\ref{tab:HFRE_ablation}, we evaluate the effect of different feature combinations and training strategies. Our full model, VLM-FO1-3B, which combines primary and auxiliary region features and keeps the primary vision encoder (QwenViT) frozen, achieves the highest average score of 67.65. In contrast, using only the auxiliary region feature (65.89) or only the primary region feature (66.15) leads to a drop in performance, demonstrating the importance of combining both semantic and perceptual information from the two vision encoders. Furthermore, we observe that fine-tuning the primary vision encoder (66.35) slightly degrades performance, suggesting that freezing the original, well-aligned vision encoder helps to preserve its valuable semantic priors. These results strongly validate the effectiveness of our HFRE design.

\subsection{Effectiveness of the Simple Feature Pyramid}
Next, we evaluate the effectiveness of introducing the SimpleFP module for the ViT-based primary vision encoder. As shown in Table~\ref{tab:simplefp_ablation}, in a controlled setting using only the primary region feature, we compare the performance with and without the SimpleFP module. The results show that removing the SimpleFP module causes a significant drop in the average score from 66.15 to 64.94. This performance gap clearly indicates that constructing a multi-scale feature pyramid from the ViT's single-scale feature map is critical for extracting high-quality, information-rich region features.

%% file: sec/6_conclusion.tex
\section{Conclusion}
In this work, we introduced VLM-FO1, a novel framework that successfully bridges the gap between the high-level reasoning of Vision-Language Models and the demands of fine-grained visual perception. By reframing object-centric tasks from a coordinate generation problem to a feature retrieval task, we circumvent a fundamental limitation of language-centric architectures. Our proposed method, featuring a plug-and-play modular design and an innovative Hybrid Fine-grained Region Encoder, effectively enhances pre-trained VLMs with state-of-the-art perception capabilities. Our extensive experiments demonstrate that VLM-FO1 achieves exceptional performance across a wide range of benchmarks spanning object grounding, region-level understanding, and complex visual reasoning, often outperforming much larger models. Crucially, these gains are achieved without compromising the base model's original general-purpose abilities, validating our training strategy and architectural design. VLM-FO1 establishes a powerful and flexible paradigm for developing the next generation of VLMs, paving the way for models with a deeper, more actionable understanding of the visual world.

%% file: sec/X_suppl.tex
\clearpage
\setcounter{page}{1}
\maketitlesupplementary

\section{Comprehensive Dataset Overview for VLM-FO1}
\begin{table*}[h!]
  \centering
  \begin{tabular}{l | l l}
    \hline
    \textbf{Stage} & \textbf{Task Type} & \textbf{Datasets} \\
    \hline
    \multirow{3}{*}{\makecell[l]{Region-Language\\Alignment}} & OD & COCO,O365\cite{shao2019objects365},V3Det\cite{wang2023v3det} \\
    \cline{2-3}
    & Grounding & GOLDG\cite{kamath2021mdetr} \\
    \cline{2-3}
    & Region Caption & Rexverse-2M\cite{jiang2024chatrex} \\
    \hline
    \multirow{8}{*}{Perception SFT} & OD & \makecell[l]{COCO,V3Det,O365,VAW\cite{pham2021learning} \\ VisDrone2019\cite{du2019visdrone},LVIS\cite{gupta2019lvis}} \\
    \cline{2-3}
    & REC & \makecell[l]{Refcoco/+/g\cite{kazemzadeh2014referitgame,mao2016generation,yu2016modeling}\\finecopsref\cite{liu2024finecops},grefcoco\cite{liu2023gres}\\CREC\cite{yu2024revisiting}} \\
    \cline{2-3}
    & Grounding & GRIT\cite{peng2023kosmos} \\
    \cline{2-3}
    & Region Caption & \makecell[l]{PACO\cite{ramanathan2023paco},VG\cite{krishna2017visual},Osprey\cite{yuan2024osprey}\\shareGPT4v\cite{chen2024sharegpt4v},Rexverse-2M} \\
    \cline{2-3}
    & \makecell[l]{Region\\Reasoning} & VisualCoT\cite{shao2024visual},HumanRef-CoT\cite{jiang2025rex} \\
    \cline{2-3}
    & Counting & \makecell[l]{COCO,LVIS,HumanRef-CoT\\CrowdHuman\cite{shao2018crowdhuman},TallyQA\cite{acharya2019tallyqa}} \\
    \cline{2-3}
    & Region QA & \makecell[l]{Osprey,VCR\cite{zellers2019recognition}\\MDVP\cite{lin2024draw},DoclayNet\cite{pfitzmann2022doclaynet}} \\
    \cline{2-3}
    & OCR & \makecell[l]{MLT2019\cite{nayef2019icdar2019},ICDAR15\cite{karatzas2015icdar}\\CurvedSynText150k\cite{liu2020abcnet}}\\
    \cline{2-3}
    & VLM-Instruction & OmChat-SFT\cite{zhao2024omchat} \\ 
    \hline
  \end{tabular}
  \caption{\textbf{A summary of the datasets used in our two-stage training process.}}
  \label{tab:training_data}
\end{table*}

This section details the comprehensive datasets utilized in our two-stage training methodology in Table~\ref{tab:training_data}.

\section{Additional Visualizations}
\label{sec:More Visualization}

\begin{figure*}[ht]
  \centering
  \includegraphics[width=1.0\textwidth]{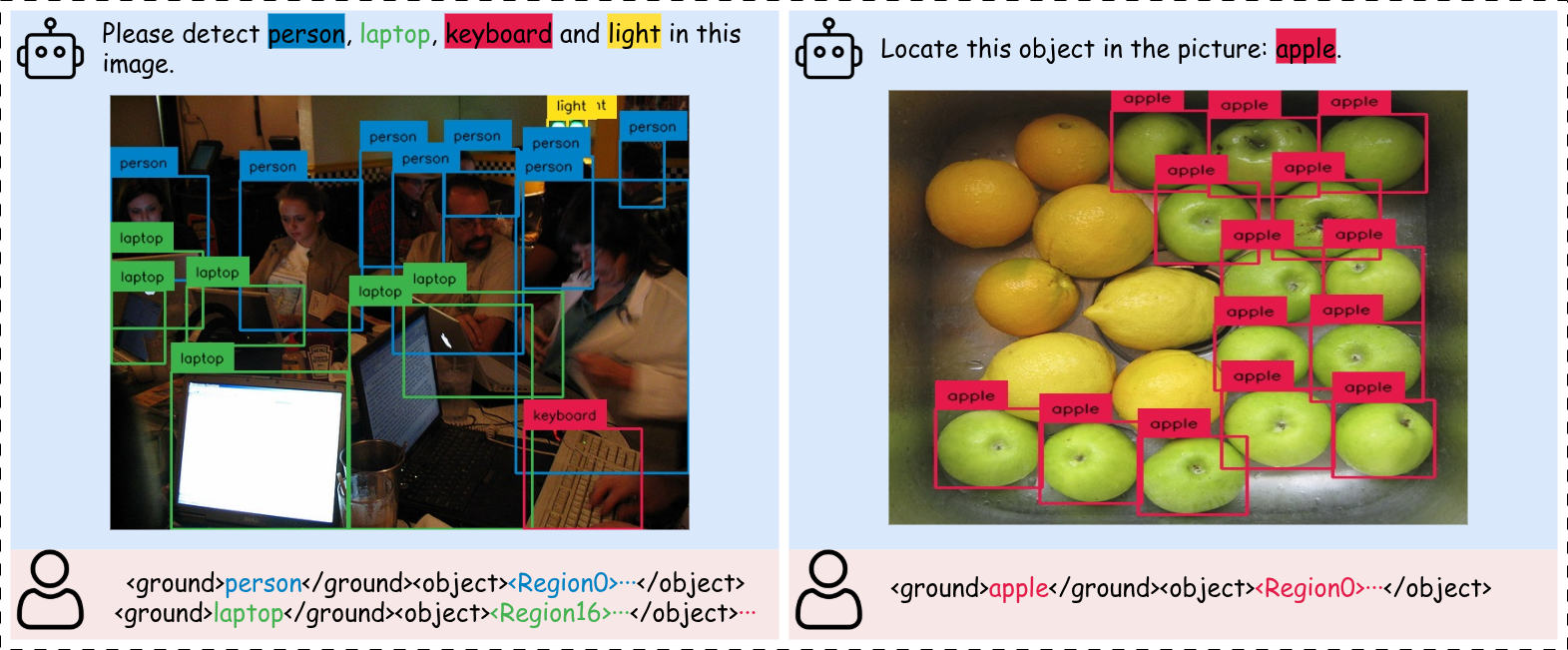}
  \caption{Visualization of VLM-FO1's perception abilities on OD task.}
  \label{fig:od_vis}
\end{figure*}

\begin{figure*}
  \centering
  \includegraphics[width=1.0\textwidth]{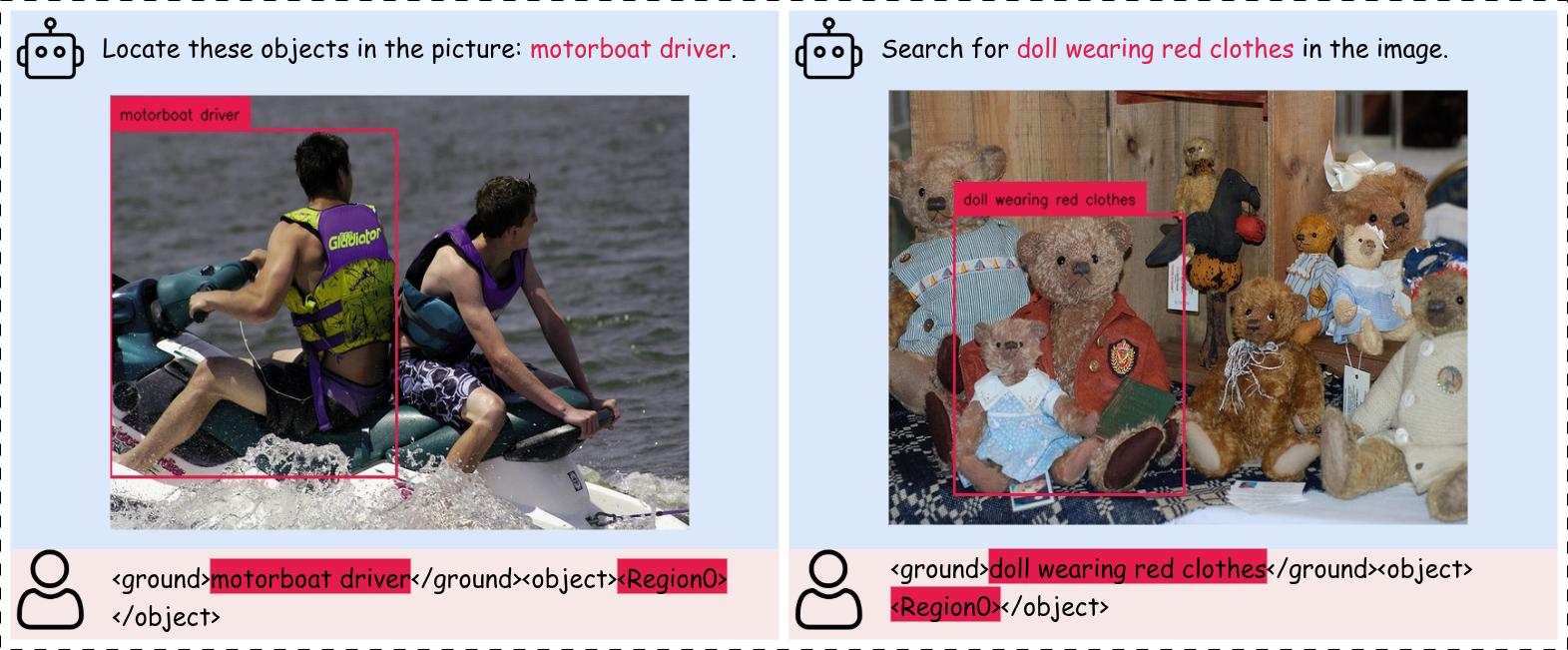}
  \caption{Visualization of VLM-FO1's perception abilities on REC task.}
  \label{fig:rec_vis}
\end{figure*}

\begin{figure*}
  \centering
  \includegraphics[width=1.0\textwidth]{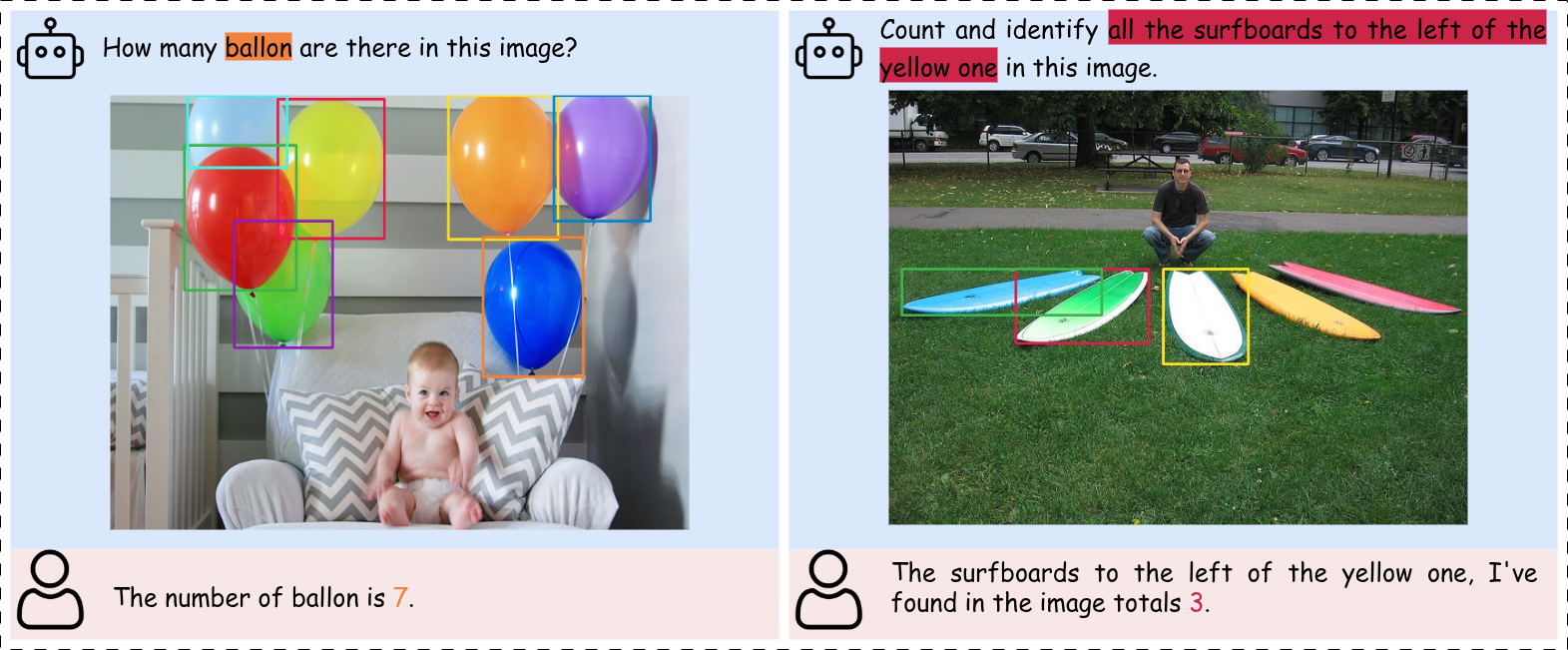}
  \caption{Visualization of VLM-FO1's perception abilities on object counting task.}
  \label{fig:counting_vis}
\end{figure*}

\begin{figure*}
  \centering
  \includegraphics[width=1.0\textwidth]{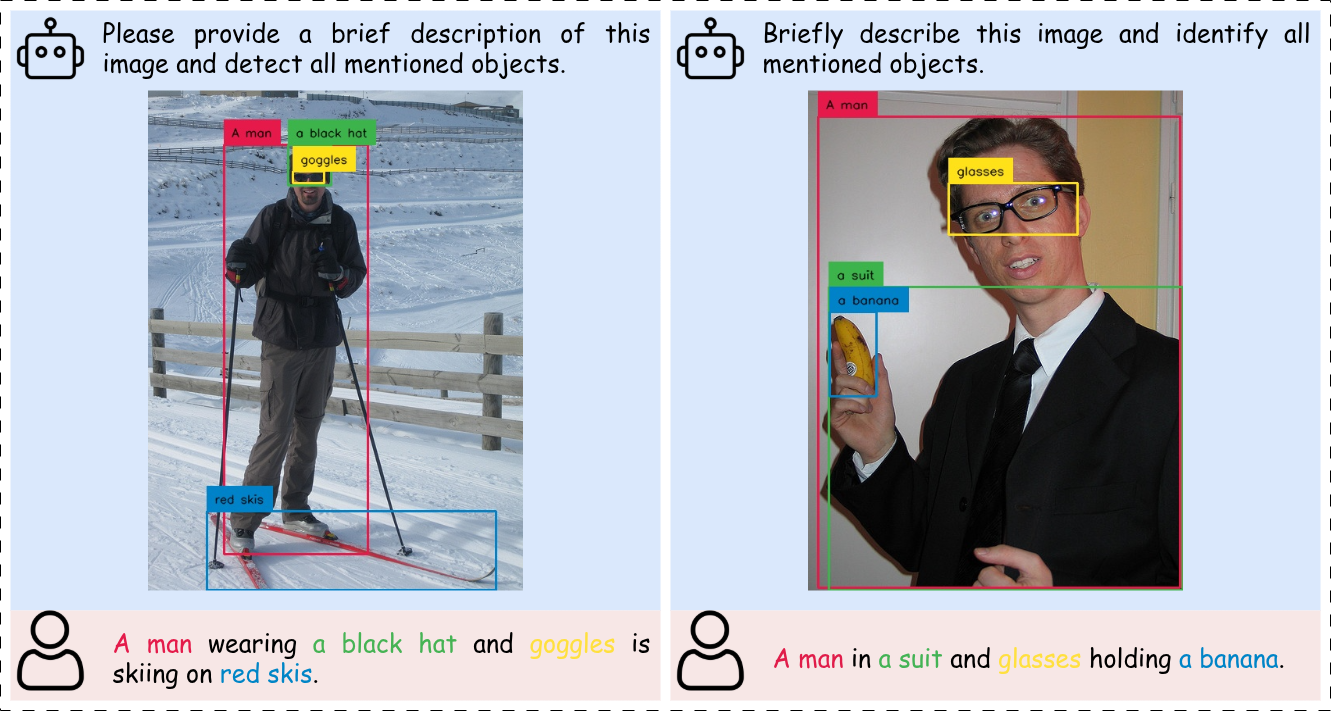}
  \caption{Visualization of VLM-FO1's perception abilities on grounding task.}
  \label{fig:grounding_vis}
\end{figure*}

\begin{figure*}
  \centering
  \includegraphics[width=1.0\textwidth]{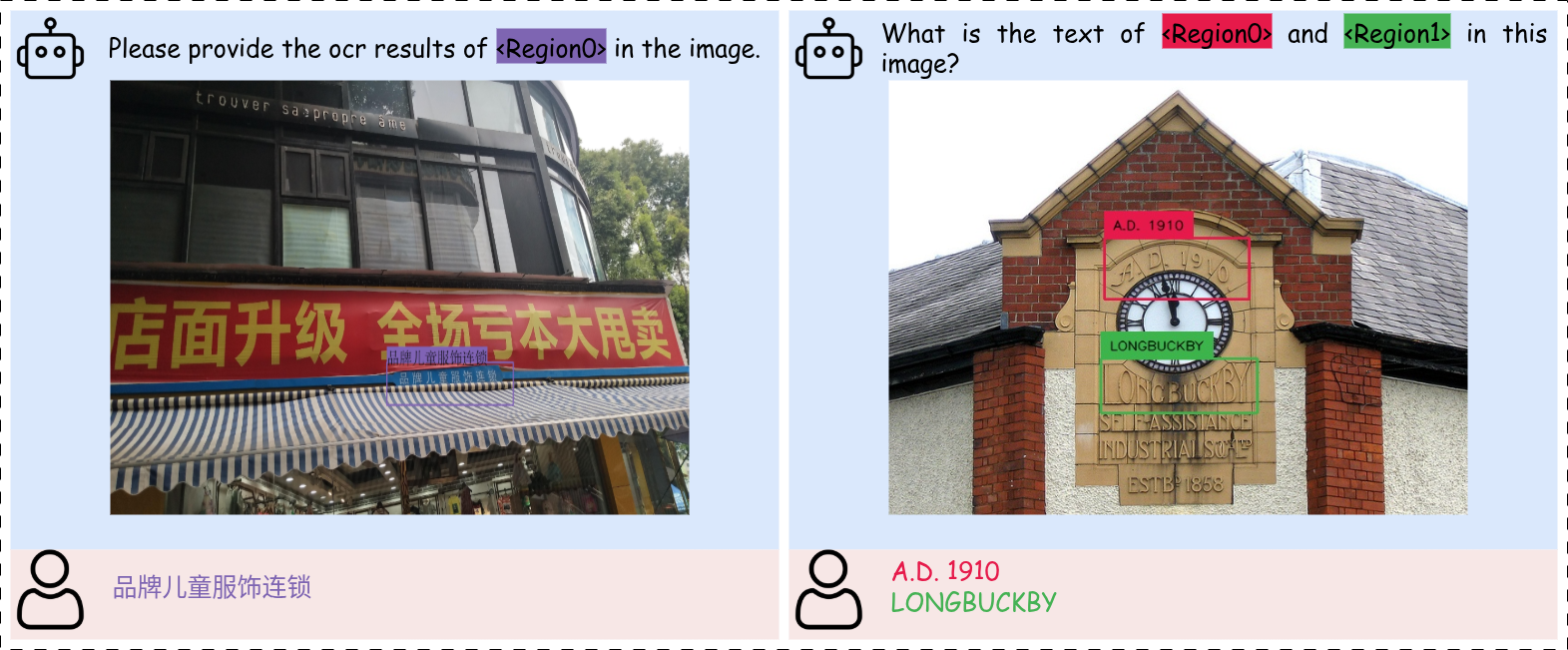}
  \caption{Visualization of VLM-FO1's perception abilities on OCR task.}
  \label{fig:ocr_vis}
\end{figure*}

\begin{figure*}
  \centering
  \includegraphics[width=1.0\textwidth]{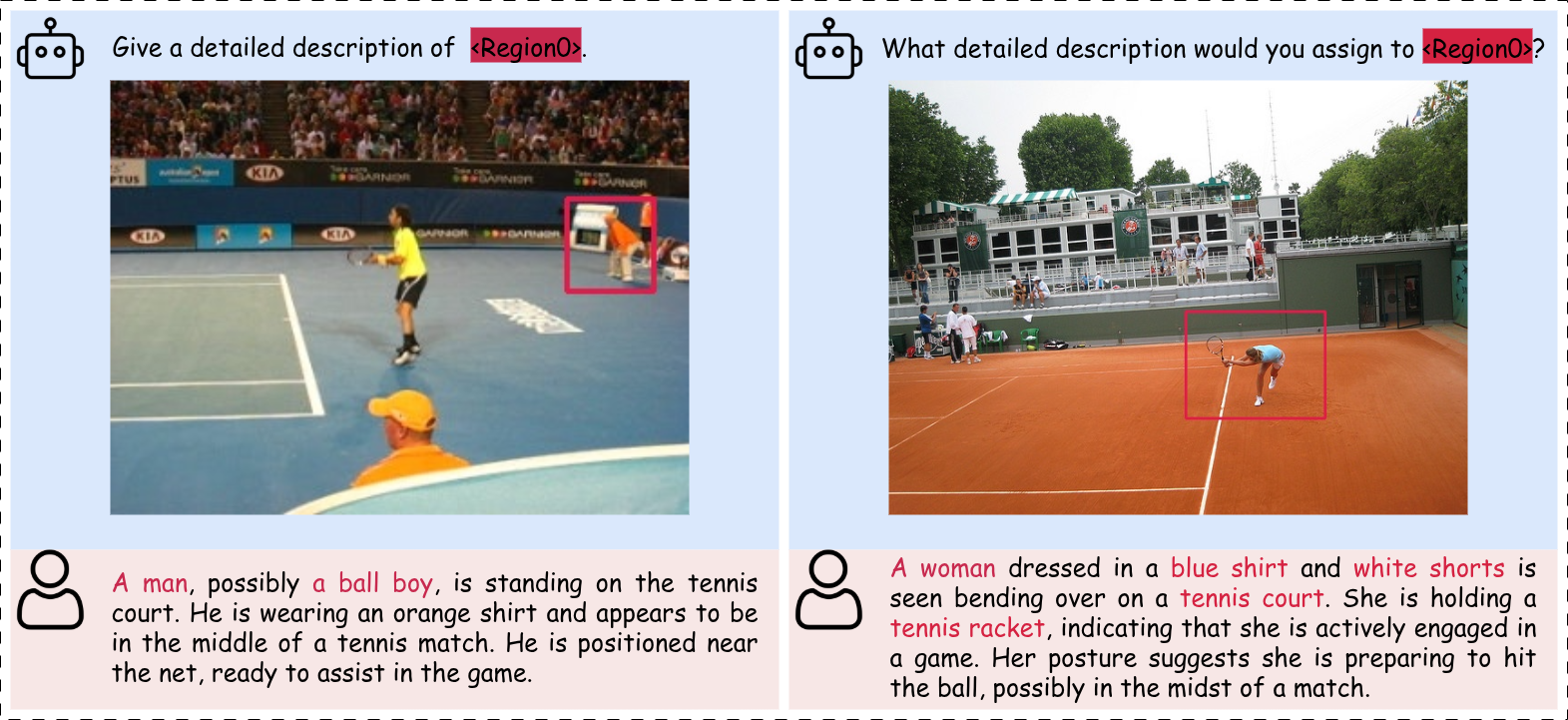}
  \caption{Visualization of VLM-FO1's perception abilities on region caption task.}
  \label{fig:regioncap_vis}
\end{figure*}

\begin{figure*}
  \centering
  \includegraphics[width=1.0\textwidth]{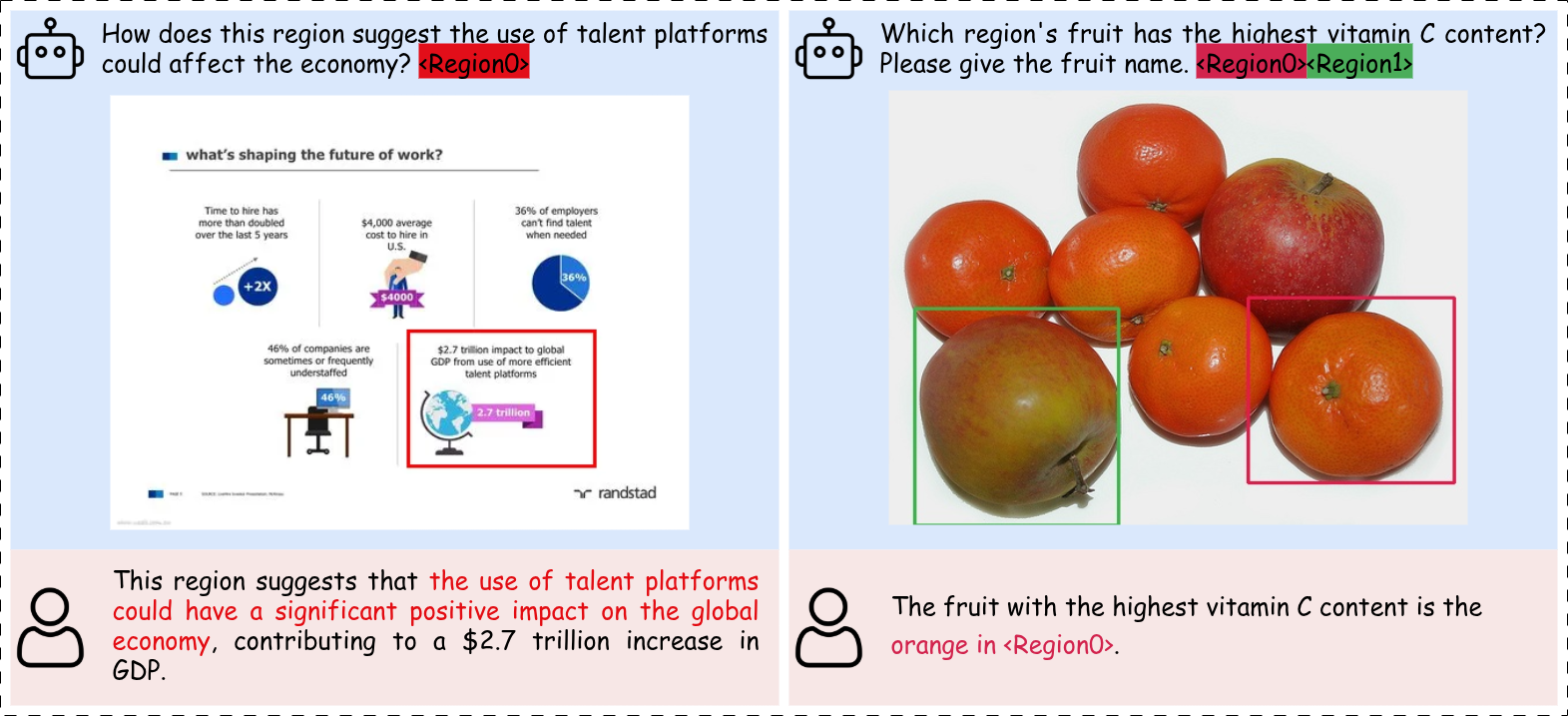}
  \caption{Visualization of VLM-FO1's perception abilities on region VQA task.}
  \label{fig:regionVQA_vis}
\end{figure*}

\begin{figure*}
  \centering
  \includegraphics[width=1.0\textwidth]{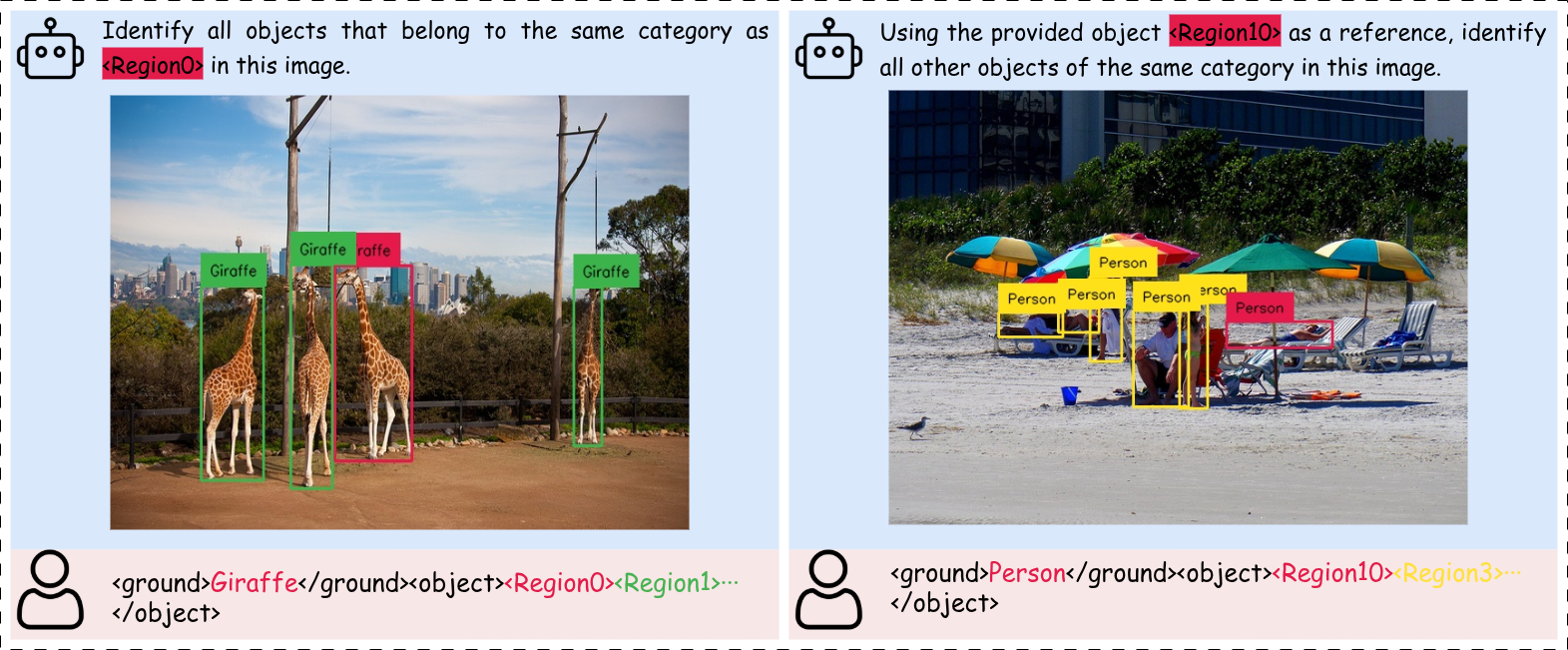}
  \caption{Visualization of VLM-FO1's perception abilities on visual prompt OD task.}
  \label{fig:vpod_vis}
\end{figure*}

\begin{figure*}
  \centering
  \includegraphics[width=1.0\textwidth]{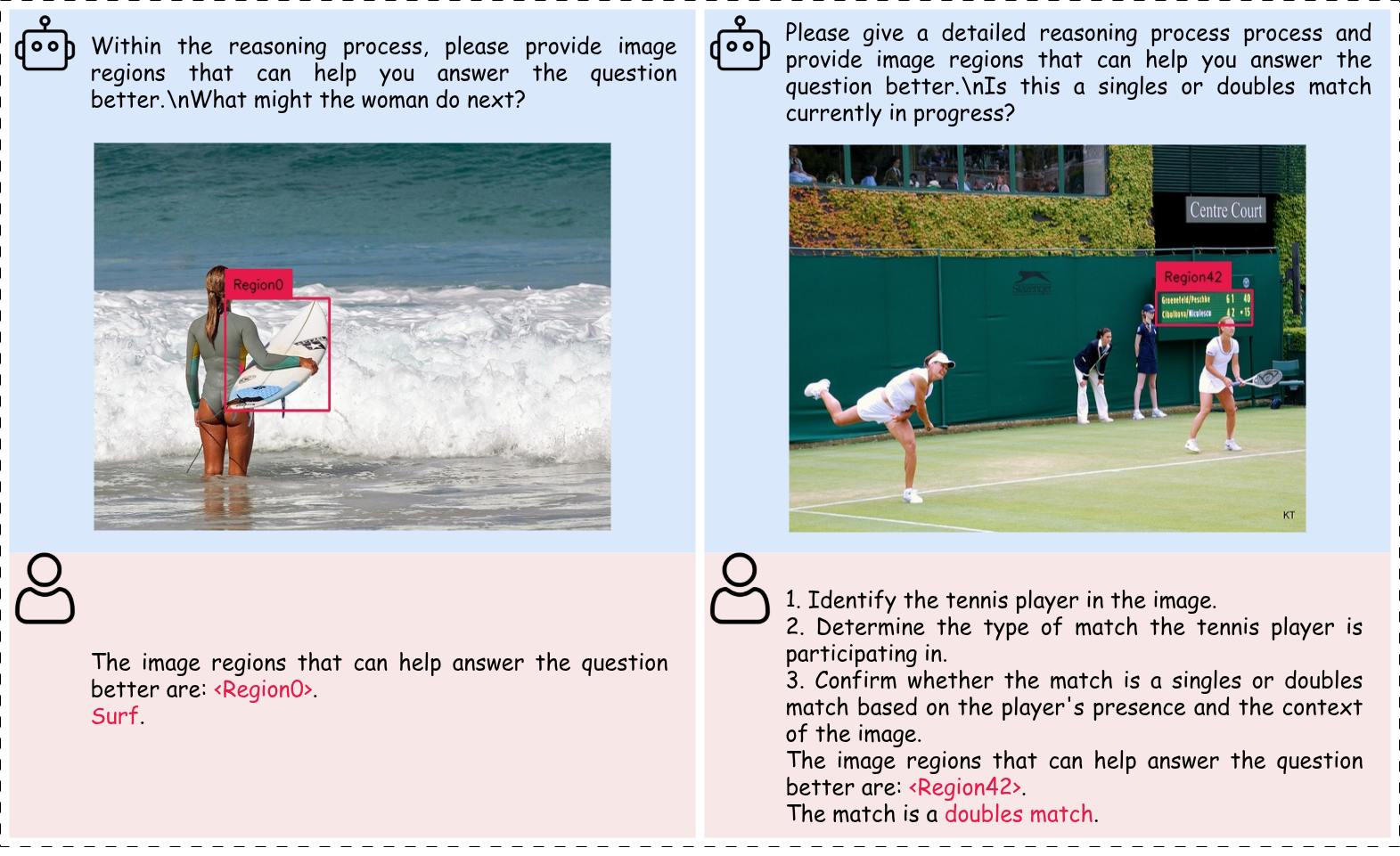}
  \caption{Visualization of VLM-FO1's perception abilities on visual region reasoning task.}
  \label{fig:cot_vis}
\end{figure*}

\begin{figure*}
  \centering
  \includegraphics[width=1.0\textwidth]{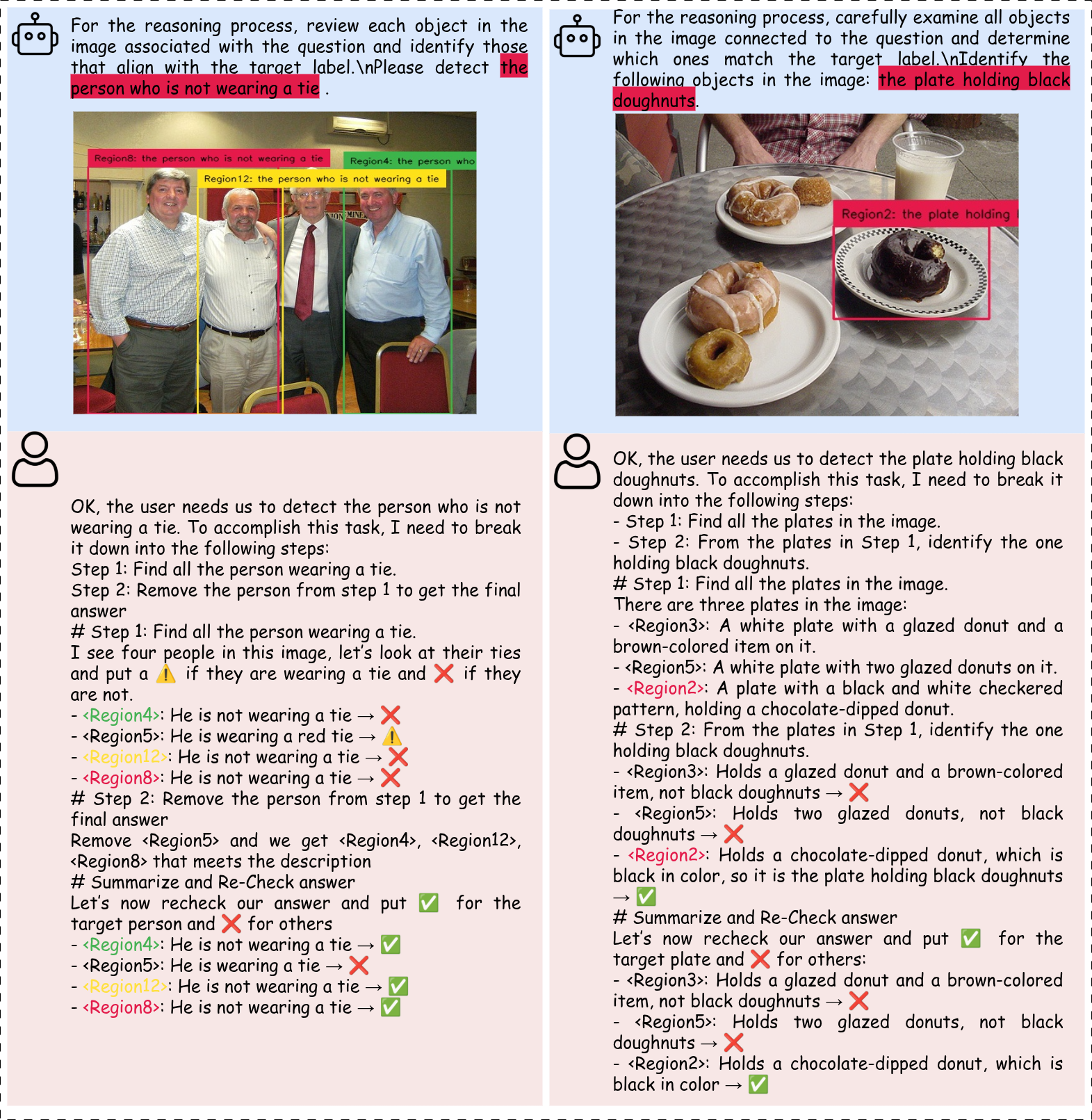}
  \caption{Visualization of VLM-FO1's perception abilities on visual region reasoning task.}
  \label{fig:cot2_vis}
\end{figure*}

This section presents a more comprehensive set of visual results to further illustrate the performance and capabilities of our VLM-FO1 model. We provide diverse examples in Figure~[\ref{fig:od_vis},\ref{fig:rec_vis},\ref{fig:counting_vis},\ref{fig:grounding_vis},\ref{fig:ocr_vis},\ref{fig:regioncap_vis},\ref{fig:regionVQA_vis},\ref{fig:vpod_vis},\ref{fig:cot_vis},\ref{fig:cot2_vis}], showcasing key aspects of our method's behavior across various scenarios, offering deeper insights beyond the examples included in the main paper.

